\documentclass[runningheads]{llncs}
\usepackage[T1]{fontenc}
\usepackage{graphicx,verbatim}
\usepackage{hyperref}
\usepackage{color}

\usepackage{booktabs}     
\usepackage{multirow}     
\usepackage{threeparttable} 
\usepackage{amsmath}
\usepackage{amssymb}
\usepackage{adjustbox}
\usepackage[table]{xcolor}
\definecolor{lightgray}{gray}{0.9}

\begin{document}
\title{Network-Aware Bilinear Tokenization for Brain Functional Connectivity Representation Learning}
\titlerunning{Network-Aware Bilinear Tokenization for Brain Functional Connectivity}

%
\author{Leo Milecki\inst{1} \and
Qingyu Hu\inst{1,2} \and
Bahram Jafrasteh \inst{1} \and
Mert R. Sabuncu\inst{1,2} \and
Qingyu Zhao\inst{1}}

%
\authorrunning{L. Milecki et al.}
%
\institute{Department of Radiology, Weill Cornell Medicine, New York, NY, USA. \and
School of Electrical and Computer Engineering, Cornell University and Cornell Tech, New York, NY, USA.
\email{lem4012@med.cornell.edu}}
  
\maketitle              
\begin{abstract}
Masked autoencoders (MAEs) have recently shown promise for self-supervised representation learning of resting-state brain functional connectivity (FC). However, a fundamental question remains unresolved: how should FC matrices be tokenized to align with the intrinsic modular organization of large-scale brain networks? Existing approaches typically adopt region-centric or graph-based schemes that treat FC as structurally homogeneous elements and overlook the large-scale network brain organization. We introduce NERVE (Network-Aware Representations of Brain Functional Connectivity via Bilinear Tokenization), a self-supervised learning framework that redefines FC tokenization by partitioning FC matrices into patches of intra- and inter-network connectivity blocks. Unlike image-based MAE, where fixed-size patches share a common tokenizer, FC patches defined by network pairs are heterogeneous in size and correspond to distinct functional roles. To resolve this problem, NERVE embeds FC patches through a novel structured bilinear factorization. This formulation preserves network identity and reduces parameter complexity from quadratic to linear scaling in the number of networks. We evaluate NERVE across three large-scale developmental cohorts (ABCD, PNC, and CCNP) for behavior and psychopathology prediction. Compared to structurally agnostic MAE variants and graph-based self-supervised baselines, the proposed network-aware formulation yields more stable and transferable representations, particularly in cross-cohort evaluation. Ablation studies confirm that the proposed bilinear network embedding and anatomically grounded parcellation are critical for performance. These findings highlight the importance of incorporating domain-specific structural priors into self-supervised learning for functional connectomics.
Code is available at: \url{https://github.com/leomlck/NERVE}


\end{abstract}

\section{Introduction}
\label{sec:intro}

Resting-state functional magnetic resonance imaging (rs-fMRI) enables the estimation of functional connectivity (FC), defined as the temporal correlation between spatially distributed brain regions. FC has become a central tool for studying individual differences in large-scale brain organization and their association with cognition, behavior, and mental health \cite{Kawahara2017BrainNetCNN:Neurodevelopment,Li2021BrainGNN:Analysis,Kan2022BrainTransformer,Wei2024AnalyzingRevealed}. 
However, extracting compact predictive representations from FC remains challenging due to its high dimensionality, low signal-to-noise ratio, and substantial inter-subject variability \cite{Woo2017BuildingNeuroimaging,Button2013PowerNeuroscience,Tiego2023PrecisionPsychopathology}.
Indeed, large-scale studies have reported that increasing model complexity does not reliably improve performance over classical approaches \cite{He2020DeepDemographics,Pervaiz2020OptimisingFMRI,Schulz2020DifferentDatasets,He2018IsIntelligence}, suggesting that more appropriate inductive biases may be required.

Masked autoencoders (MAE)~\cite{He2021MaskedLearners} provide a principled framework for representation learning by partitioning the input into units, embedding each unit into a token representation, masking a subset of tokens, and reconstructing the masked content from the visible ones. In computer vision, these units correspond to spatial image patches that naturally align with local structure. When adapted to FC, however, the notion of a “patch” lacks a canonical definition, as the arrangement of regions in an FC matrix is largely arbitrary and does not necessarily reflect spatial or functional locality. 
Defining appropriate patch units and their corresponding token embeddings thus becomes central for applying MAE to FC.
Existing approaches for FC adopt heuristic tokenization schemes prior to masking \cite{Dong2024Brain-JEPA:Masking,Yang2024BrainMass:Learning,Caro2024BrainLM:Recordings,Ma2025RS-MAE:FMRI,Gao20263DData}. BrainMass~\cite{Yang2024BrainMass:Learning}, for example, treats individual regions as units and masks randomly selected rows of the FC matrix, while RS-MAE~\cite{Ma2025RS-MAE:FMRI} masks grouped regions inspired by spatiotemporal strategies. Graph-based methods similarly operate on node-level embeddings, where masking is applied to node tokens ~\cite{Farahani2019ApplicationReview,Hou2022GraphMAE:Autoencoders,Peng2023GATE:Analysis,Wen2023GraphAnalysis}. 
Despite promising results, a fundamental question remains: how should FC matrices be partitioned into patches and embedded into tokens to align the learned representations with the modular, network-level organization of brain FC? 

\begin{figure}[!t]
\includegraphics[width=1.0\textwidth]{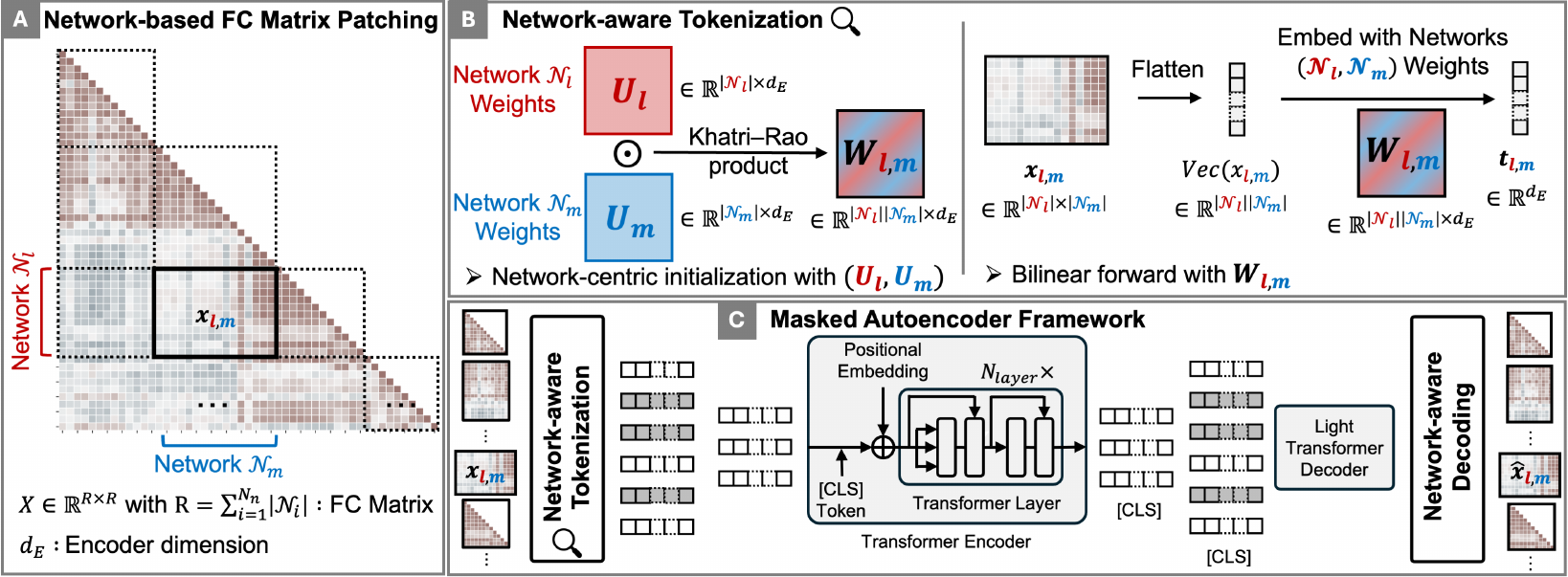}
\caption{\textbf{Overview of NERVE.}
\textbf{A.} The functional connectivity (FC) matrix is partitioned into patches defined by pairs of functional brain networks.
\textbf{B. Network-aware Bilinear Tokenization.} Each functional network is assigned learnable network-specific weights at initialization, and patch tokens are computed through structured bilinear interactions between network weights during forward.
\textbf{C. MAE Framework.} We apply a standard MAE framework to the proposed network-aware tokens, thereby introducing a functionally informed inductive bias over connectivity structure.
}
\label{fig:overview}
\end{figure}

Our central insight is that the conceptual analog of spatially neighboring pixels in an image is groups of brain regions that share similar functional dynamics, e.g., regions organized into large-scale functional networks~\cite{ThomasYeo2011TheConnectivity}. Under this view, the natural counterpart of an image patch is a connectivity block defined by intra- or inter-network interactions. 
However, a key challenge arises: FC patches defined by pairs of networks vary in dimensionality, precluding the use of a shared patch encoder. Moreover, each network carries distinct functional roles, suggesting that tokenization should preserve network identity rather than collapse all patches into a homogeneous representation space. To address this, we introduce a novel and parameter-efficient \emph{bilinear tokenization} scheme. Instead of learning independent embeddings for each network-pair patch, we learn network-specific region embeddings and model inter-network connectivity via bilinear interactions. This factorization replaces quadratic growth in patch-specific parameters with linear scaling in the number of networks, while explicitly encoding network identity and structured intra- and inter-network interactions.

Integrating this design into an MAE framework, we introduce \textbf{NERVE} (Network-Aware Representations of Brain Functional Connectivity via Bilinear Tokenization), a self-supervised approach tailored to the modular organization of brain FC. We evaluate NERVE across three large-scale adolescent neuroimaging cohorts on the challenging task of predicting behavioral and psychopathology scores. Our results demonstrate that NERVE learns more informative and transferable FC representations, outperforming alternative tokenization strategies and existing self-supervised learning (SSL) approaches.

\section{Methods}
\label{sec:meth}

Let $\mathcal{D} = \{X^{(i)}\}_{i=1}^{N}$ denote FC matrices of $N$ participants, where each $X^{(i)} \in \mathbb{R}^{R \times R}$ is a correlation matrix constructed from functional time series across $R$ brain regions. 
We aim to learn structured and transferable representations of $X$ in a self-supervised manner. 
To this end, we adopt a transformer-based MAE framework, which partitions $X$ into patches, encodes each patch into a token, randomly masks a subset of tokens, and reconstructs the masked content from the visible tokens.

\textbf{Network-based Patching.} The way data are partitioned into patches determines what structure the model can exploit through masking and reconstruction.
While image-based MAE relies on a natural decomposition into fixed-size image patches, FC matrices lack a canonical patching scheme, making tokenization a non-trivial design choice. Rather than treating rows of $X$ as patches \cite{Dong2024Brain-JEPA:Masking,Yang2024BrainMass:Learning,Caro2024BrainLM:Recordings}, we observe that the conceptual analog of neighboring pixels in images is groups of brain regions that share similar functional dynamics. This organization is captured by large-scale functional networks~\cite{ThomasYeo2011TheConnectivity}. Therefore, we propose to reorganize $X$ by grouping the $R$ regions into $N_n$ established functional networks $\{\mathcal{N}_1, \dots, \mathcal{N}_{N_n}\}$ (e.g., Visual, Default, Dorsal Attention). For each network pair $(l,m)$ with $l \leq m$, we define a connectivity block:
$x_{l,m} \in \mathbb{R}^{|\mathcal{N}_l| \times |\mathcal{N}_m|}$
representing intra- ($l=m$) or inter-network ($l < m$) connectivity (Fig.~\ref{fig:overview}A). These connectivity patches are then treated as the basic units for masking and reconstruction. The total number of patches is:
$N_{\text{patch}} = \frac{N_n (N_n + 1)}{2}.$

\textbf{Shared vs. Patch-specific Tokenization.}
In image-based MAE, patches are fixed-size and interchangeable, enabling a shared projection for tokenization. Here, network-defined patches vary in size and correspond to specific network interactions, making them structurally and semantically distinct. Embedding these irregular and network-specific patches, therefore, requires a dedicated network-aware tokenization strategy.
A straightforward adaptation of image-based MAE to FC would flatten each patch $x_{l,m}$ and project it through a linear transformation (denoted as \textit{shared} linear):
$t_{l,m} = W^{\top} \,\mathrm{vec}(x_{l,m}),$
where $W \in \mathbb{R}^{S_{\text{max}} \times d_{E}}$ with $d_{E}$ the encoder embedding dimension, and \( S_{\text{max}} = \max_{l,m} |\mathcal{N}_l| \cdot |\mathcal{N}_m| \) the maximum flattened patch size.
Because patches have different sizes, zero-padding to the largest patch size $S_{\max}$ is required. While parameter-efficient, this shared projection enforces a common representation across semantically distinct network interactions and ignores the structural heterogeneity of FC.
An alternative is to assign a distinct projection layer to each network pair (denoted as patch-\textit{specific} linear):
$t_{l,m} = W_{l,m}^{\top} \,\mathrm{vec}(x_{l,m}),$
where $W_{l,m} \in \mathbb{R}^{(|\mathcal{N}_l| |\mathcal{N}_m|) \times d_{E}}$.
Although this allows patch-specific (network pairs) modeling, it introduces a quadratic growth in parameters with respect to the number of networks $N_n$, which quickly becomes impractical and risks overfitting.

\textbf{Bilinear Tokenization.}
To preserve network specificity while maintaining parameter efficiency, we propose a bilinear network-aware tokenization.
Each functional network $\mathcal{N}_l$ is assigned a learnable matrix $U_l \in \mathbb{R}^{|\mathcal{N}_l| \times d_E}$, where each column represents a network-specific embedding dimension.
For an FC patch $x_{l,m} \in \mathbb{R}^{|\mathcal{N}_l| \times |\mathcal{N}_m|}$ between networks $l$ and $m$, we construct the corresponding tokenizer via a column-wise Kronecker (Khatri--Rao)~\cite{Khatri1968SolutionsDistributions} product (Fig.~\ref{fig:overview}B):
\[
W_{l,m} = U_l \odot U_m
\;\in\;
\mathbb{R}^{(|\mathcal{N}_l||\mathcal{N}_m|) \times d_E},
\quad
t_{l,m}
=
W_{l,m}^\top \mathrm{vec}(x_{l,m})
\;\in\;
\mathbb{R}^{d_E}.
\]
where $\odot$ denotes the Khatri--Rao product defined elementwise as
$
[W_{l,m}]_{(i,j),k}
=
[U_l]_{i,k}\,[U_m]_{j,k}.
$
Conceptually, instead of learning an independent projection for each network pair, we learn network-level region embeddings and model inter-network connectivity as bilinear interactions between them.
The contribution of a connection between region $i \in \mathcal{N}_l$ and region $j \in \mathcal{N}_m$ to embedding dimension $k$ is given by the product $[U_l]_{i,k}[U_m]_{j,k}$, capturing structured interactions while sharing parameters across networks.
This low-rank factorization of patch-specific projections replaces quadratic growth in patch-specific parameters with a linear scaling in the number of networks, while explicitly encoding network identity and structured intra- and inter-network interactions.
Specifically, if we assume the $R$ brain regions are approximately evenly distributed across the $N_n$ networks, we can approximate $S_{\text{max}} \approx (R / N_n)^2$. Under this assumption, the parameter complexities for \textit{shared} linear, patch-\textit{specific} linear, and \textit{bilinear} embedding are: $\mathcal{O}\left(\frac{R^2}{N_n^2} \times d_{\text{E}}\right)$,  $\mathcal{O}\left(R^2 \times d_{\text{E}}\right)$, and $\mathcal{O}\left(R \times d_{\text{E}}\right)$, respectively, highlighting the trade-off between parameter efficiency and expressivity by shifting complexity from universal, patch-specific to network-centric representations.

\textbf{MAE for FC.}
Given the network-aware token sequence $T \in \mathbb{R}^{N_{\text{patch}} \times d_{E}}$, 
a learnable CLS token is prepended, and learnable positional embeddings are added to encode the identity of each network pair.
For each subject, a fixed proportion of patch tokens is randomly masked, and only the visible tokens are processed by a transformer encoder, producing contextualized representations.
To reconstruct masked connectivity, mask tokens are first inserted at masked positions, and the full token sequence is then processed by a lightweight transformer decoder. The decoded tokens are then projected back to the FC patch space using a bilinear decoding layer consistent with the tokenization scheme, yielding reconstructed patches $\hat{x}_{l,m}$ (Fig.~\ref{fig:overview}C).
The reconstruction objective is:
$\mathcal{L}_{\text{recon}} = \frac{1}{|\mathcal{M}|} \sum_{(l,m) \in \mathcal{M}} \left\| \hat{x}_{l,m} - x_{l,m} \right\|_2^2$.

\textbf{Implementation Details.} We use Schaefer's 17-network parcellation ($R=400$, $N_n=17$, $N_{\text{patch}}=153$)~\cite{Schaefer2018Local-GlobalMRI} with a masking ratio of $0.5$. The encoder uses $4$ layers, $4$ heads, and $d_{E}=256$. The decoder has $1$ layer, $2$ heads with $d_{D}=64$.
Models are trained for 4,000 epochs using AdamW with a learning rate of $1e^{-2}$, weight decay of $1e^{-2}$, cosine scheduling with linear warmup for 400 epochs, batch size of 1,024, and mixed precision using Pytorch on a single NVIDIA L40 GPU. 
We perform preliminary hyperparameter tuning on the training data for the masking ratio, transformer encoder, and decoder length, width, and learning rate. For brevity, we report results using the best-performing configuration and focus ablations on methodological design choices central to our contribution.

\section{Experiments and Results}
\label{sec:exp}

\textbf{Datasets \& Downstream Evaluation.}
We evaluate our framework on three large-scale developmental rs-fMRI cohorts: ABCD (N=1,791, age 9–10, 53.8\% F)~\cite{Garavan2018RecruitingProcedures}, PNC (N=1,416, age 8–23, 53.6\% F)~\cite{Satterthwaite2014NeuroimagingCohort}, and CCNP (N=178, age 6–17, 52.3\% F)~\cite{Liu2021ChineseCohort}. 
For ABCD, we directly use the FC matrices preprocessed and released by CBIG~\cite{Ooi2022ComparisonMRI}, which contain a subset of the full ABCD cohort with the highest image quality. We preprocess PNC and CCNP data using the publicly available DeepPrep pipeline~\cite{Ren2025DeepPrep:Learning}, which includes motion correction, nuisance regression, and temporal and spatial filtering. All FC matrices are defined by the $R=400$ regions Schaefer atlas~\cite{Schaefer2018Local-GlobalMRI}.
We train the MAE on the FC matrices from the combined ABCD+PNC datasets, treating CCNP as out-of-domain samples, and apply the trained model to extract FC representations for all three datasets.
Then, within each dataset, we use the representations to predict behavioral phenotypes.
For ABCD, we predict the \textit{Internalizing}, \textit{Externalizing}, and \textit{Total} scores from the Child Behavior Checklist (CBCL)~\cite{Achenbach1978TheEfforts}. 
For PNC and CCNP, we predict the Reproducible Brain Charts (RBC)~\cite{Hoffmann2024AnAdolescents} harmonized \textit{Internalizing}, \textit{Externalizing}, and general psychopathology (\textit{p-factor}) scores.
Kernal Ridge Regression (KRR) is adopted as the downstream regression model following prior large-scale evaluations demonstrating its strong and stable performance for behavioral prediction compared to more complex neural network-based probes~\cite{He2020DeepDemographics}.
KRR is assessed using stratified 10-fold cross-validation within each dataset, with age and sex effects removed via linear regressions fit on the training folds.
Finally, we report the Pearson correlation between the true and predicted behavioral scores concatenated across testing folds. Uncertainty is quantified via bootstrap resampling over subjects (1,000 iterations) to estimate 95\% confidence intervals (CI).
The statistical significance of performance differences between the top two methods was evaluated using two-sided paired bootstrap tests on subject-level out-of-fold predictions (1,000 iterations) for each behavioral variable.

\begin{table*}[!t]
\centering
\caption{\textbf{Behavioral prediction performance across developmental cohorts.}
We report Pearson correlation $r_{\pm\Delta}$ with bootstrap 95\% CI half-width $\Delta$ from 10-fold cross-validation. n.: negative values, OOD: Out-of-domain.}
\label{tab:main}
\begin{adjustbox}{width=1.0\textwidth}
\begin{tabular}{l|ccc|ccc||ccc}
\toprule
\textbf{Dataset $\rightarrow$} 
& \multicolumn{3}{c|}{\textbf{ABCD}} 
& \multicolumn{3}{c||}{\textbf{PNC}} 
& \multicolumn{3}{c}{\textbf{CCNP} (OOD)} \\
\cmidrule(lr){2-4} \cmidrule(lr){5-7} \cmidrule(lr){8-10}
\textbf{Method $\downarrow$} 
& \textit{Int.} & \textit{Ext.} & \textit{Total} 
& \textit{Int.} & \textit{Ext.} & \textit{p-factor} 
& \textit{Int.} & \textit{Ext.} & \textit{p-factor}  \\
\midrule

BrainNetCNN
& $.05_{\pm .04}$ & \underline{.08}$_{\pm .04}$ & \underline{.11}$_{\pm .04}$ 
& $.07_{\pm .03}$ & $.08_{\pm .04}$ & $.05_{\pm .03}$ 
& \underline{.15}$_{\pm .12}$ & \underline{.10}$_{\pm .13}$ & $.01_{\pm .12}$ \\

BrainGNN
& $.05_{\pm .04}$ & \underline{.08}$_{\pm .05}$ & $.06_{\pm .04}$ 
& $.08_{\pm .04}$ & $.04_{\pm .04}$ & $.06_{\pm .04}$ 
& \textbf{.17}$_{\pm .13}$ & $.16_{\pm .16}$ & $.04_{\pm .10}$ \\

BrainNetTF 
& -- & -- & --  
& $.08_{\pm .04}$ & $.07_{\pm .05}$ & $.05_{\pm .04}$ 
& $.01_{\pm .09}$ & $.21_{\pm .14}$ & $.05_{\pm .09}$ \\

\midrule
GraphMAE
& \underline{.06}$_{\pm .05}$ & $.06_{\pm .04}$ & $.06_{\pm .04}$ 
& \underline{.09}$_{\pm .05}$ & $.08_{\pm .05}$ & $.09_{\pm .05}$ 
& $.07_{\pm .16}$ & \underline{.22}$_{\pm .16}$ & \underline{.12}$_{\pm .13}$ \\

GATE
& -- & -- & --  
& $.07_{\pm .06}$ & $.08_{\pm .05}$ & $.09_{\pm .04}$ 
& $.04_{\pm .09}$ & $.14_{\pm .14}$ & $.11_{\pm .08}$ \\

BrainGSLs 
& $.04_{\pm .05}$ & $.05_{\pm .05}$ & $.05_{\pm .05}$ 
& $.05_{\pm .05}$ & $.08_{\pm .06}$ & $.04_{\pm .05}$ 
& n. & \underline{.22}$_{\pm .15}$ & n. \\

BrainMass 
& \underline{.06}$_{\pm .05}$ & \underline{.08}$_{\pm .05}$ & $.01_{\pm .04}$ 
& $.03_{\pm .06}$ & \underline{.09}$_{\pm .06}$ & \underline{.10}$_{\pm .05}$ 
& $.04_{\pm .13}$ & $.14_{\pm .14}$ & \textbf{.13}$_{\pm .16}$ \\

\textbf{NERVE} 
& \cellcolor{lightgray}\textbf{.11}$_{\pm .06}$ & \textbf{.09}$_{\pm .05}$ & \textbf{.13}$_{\pm .04}$ 
& \cellcolor{lightgray}\textbf{.14}$_{\pm .05}$ & \cellcolor{lightgray}\textbf{.13}$_{\pm .05}$ & \textbf{.12}$_{\pm .05}$ 
& $.08_{\pm .14}$ & \cellcolor{lightgray}\textbf{.33}$_{\pm .14}$ & \textbf{.13}$_{\pm .15}$ \\

\bottomrule
\multicolumn{10}{l}{{Gray shading: $p<0.05$, two-sided paired bootstrap test between top-\textbf{1} and \underline{2} methods.}}
\end{tabular}
\end{adjustbox}
\end{table*}

\textbf{Comparison Methods.}
We benchmark NERVE against established methods for FC encoding, restricting comparisons to approaches operating on FC matrices to ensure fair and interpretable evaluation. Compared encoders include BrainMass~\cite{Yang2024BrainMass:Learning}, a foundation-style model combining temporal masking and contrastive learning, and graph-based SSL methods: GraphMAE~\cite{Hou2022GraphMAE:Autoencoders}, BrainGSLs~\cite{Wen2023GraphAnalysis}, and GATE~\cite{Peng2023GATE:Analysis}, which apply node-masked graph encoders with SSL objectives.  
All models are assessed using the same protocol: training encoders on ABCD+PNC and evaluating downstream KRR within each dataset via 10-fold cross-validation.
To put the prediction performance in the context of fully supervised, task-specific learning capacity, we also evaluate state-of-the-art supervised FC-based prediction models, which include
BrainNetCNN~\cite{Kawahara2017BrainNetCNN:Neurodevelopment}, 
BrainGNN~\cite{Li2021BrainGNN:Analysis}, 
and BrainNetTF~\cite{Kan2022BrainTransformer}. These supervised models are trained and evaluated end-to-end within each dataset by 10-fold cross-validation, as behavioral targets derive from different assessment instruments and are not strictly interchangeable across cohorts.
GATE and BrainNetTF require BOLD time series preprocessing and are therefore evaluated only on PNC and CCNP, as only FC matrices are available for ABCD in the CBIG release.

\textbf{Main Results.}
Past rs-fMRI studies have consistently shown that continuous behavioral prediction from FC is intrinsically challenging, with state-of-the-art models typically achieving modest effect sizes (often $r<0.15$) even in large cohorts \cite{Ooi2022ComparisonMRI,He2020DeepDemographics}. In this context, the representations learned by NERVE demonstrate strong and stable predictive power across datasets (Table~\ref{tab:main}).
Specifically, in in-domain evaluations (ABCD, PNC), NERVE achieves the highest or tied-highest performance across all behavioral targets, supporting the advantage of incorporating network-level structure over structurally agnostic SSL baselines and supervised FC architectures trained within a single dataset. Among self-supervised baselines, GraphMAE and BrainMass show competitive performance on specific targets but exhibit higher variability across targets.
In the out-of-domain CCNP cohort, NERVE exhibits strong cross-cohort generalization, achieving the highest observed correlation for externalizing symptoms ($r=.33$) and competitive performance for the general psychopathology (p-factor) score. These results further support the role of network-aware tokenization in enhancing representation stability beyond the training distribution.
The only target where NERVE does not achieve top performance is the internalizing phenotype in the CCNP cohort, where supervised models trained specifically on CCNP obtain higher correlations. 

\begin{table*}[!t]
\centering
\caption{\textbf{Tokenization ablation.}
We report Pearson correlation $r_{\pm\Delta}$ with bootstrap 95\% CI half-width $\Delta$ from 10-fold cross-validation. n. indicates negative values.}
\label{tab:arch_ablation}
\begin{adjustbox}{width=1.0\textwidth}
\begin{tabular}{l|ccc|ccc||ccc}
\toprule
\textbf{Dataset $\rightarrow$} 
& \multicolumn{3}{c|}{\textbf{ABCD}} 
& \multicolumn{3}{c||}{\textbf{PNC}} 
& \multicolumn{3}{c}{\textbf{CCNP}} \\
\cmidrule(lr){2-4} \cmidrule(lr){5-7} \cmidrule(lr){8-10}
\textbf{Method $\downarrow$} 
& \textit{Int.} & \textit{Ext.} & \textit{Total} 
& \textit{Int.} & \textit{Ext.} & \textit{p-factor} 
& \textit{Int.} & \textit{Ext.} & \textit{p-factor}  \\
\midrule

{\textit{shared} linear} 
& \underline{.06}$_{\pm .06}$ & $.06_{\pm .05}$ & $.02_{\pm .05}$ 
& \underline{.10}$_{\pm .05}$ & $.11_{\pm .05}$ & \underline{.10}$_{\pm .06}$ 
& \underline{.06}$_{\pm .10}$ & \underline{.31}$_{\pm .15}$ & \underline{.12}$_{\pm .15}$ \\

{\textit{specific} linear} 
& $.03_{\pm .08}$ & \underline{.08}$_{\pm .05}$ & \underline{.06}$_{\pm .05}$ 
& \textbf{.14}$_{\pm .06}$ & \underline{.12}$_{\pm .05}$ & $.09_{\pm .06}$ 
& $.01_{\pm .11}$ & $.03_{\pm .13}$ & n. \\

\textbf{\textit{bilinear}} 
& \textbf{.11}$_{\pm .09}$ & \textbf{.09}$_{\pm .05}$ & \textbf{.13}$_{\pm .04}$ 
& \textbf{.14}$_{\pm .05}$ & \textbf{.13}$_{\pm .05}$ & \textbf{.12}$_{\pm .05}$ 
& \textbf{.08}$_{\pm .14}$ & \textbf{.33}$_{\pm .14}$ & \textbf{.13}$_{\pm .15}$ \\

\bottomrule
\end{tabular}
\end{adjustbox}
\end{table*}
\begin{table*}[!t]
\footnotesize
\centering
\caption{\textbf{Parcellation and network structure ablation.}
We report Pearson $r$ on 10-fold cross-validation. For the \textit{Permutation} baselines, we report mean $\pm$ std $r$ over 100 random permutations of region-to-network assignment.}
\label{tab:parcellation_ablation}
\begin{adjustbox}{width=1.0\textwidth}
\begin{tabular}{l|ccc|ccc||ccc}
\toprule
\textbf{Dataset $\rightarrow$} 
& \multicolumn{3}{c|}{\textbf{ABCD}} 
& \multicolumn{3}{c||}{\textbf{PNC}} 
& \multicolumn{3}{c}{\textbf{CCNP}} \\
\cmidrule(lr){2-4} \cmidrule(lr){5-7} \cmidrule(lr){8-10}
\textbf{Parcellation $\downarrow$} 
& \textit{Int.} & \textit{Ext.} & \textit{Total} 
& \textit{Int.} & \textit{Ext.} & \textit{p-factor} 
& \textit{Int.} & \textit{Ext.} & \textit{p-factor}  \\
\midrule

Permutation 
& .04$\pm$.02 & .07$\pm$.02 & .07$\pm$.03
& .09$\pm$.02 & .10$\pm$.03 & .09$\pm$.01
& -.04$\pm$.09 & .19$\pm$.05 & .00$\pm$.06 \\

\midrule
$16 \times 16$ \textit{shared}
& .02 & .02 & .05 
& .05 & .09 & \textbf{.11$^{\dagger}$} 
& -.03 & .11 & .01 \\

$16 \times 16$ \textit{specific}
& .02 & \textbf{.10$^{\dagger}$} & .06 
& .04 & .09 & .08 
& -.02 & \textbf{.25$^{\dagger}$} & \textbf{.11$^{\dagger}$} \\

$16 \times 16$ \textit{bilinear}
& .01 & .08 & .06 
& .05 & .10 & .08 
& .00 & .11 & -.05 \\

\midrule
7-network
& .05 & \textbf{.09$^{\dagger}$} & \textbf{.09$^{\dagger}$}
& .07 & .10 & .07 
& -.01 & .22 & \textbf{.07$^{\dagger}$}\\

17-network
& \textbf{.11$^{\dagger}$} & \textbf{.09$^{\dagger}$} & \textbf{.13$^{\dagger}$} 
& \textbf{.14$^{\dagger}$} & \textbf{.13$^{\dagger}$} & \textbf{.12$^{\dagger}$} 
& \textbf{.08$^{\dagger}$} & \textbf{.33$^{\dagger}$} & \textbf{.13$^{\dagger}$} \\

\bottomrule
\multicolumn{10}{l}{\textbf{Bold$^{\dagger}$}: $p < 0.05$ by one-sided permutation test compared to \textit{Permutation} baselines.}
\end{tabular}
\end{adjustbox}
\end{table*}

\textbf{Tokenization Ablation.}
We compare three tokenization strategies within the NERVE framework:
i) \textit{shared} linear,
ii) patch-\textit{specific} linear, and
iii) the proposed \textit{bilinear} network-aware formulation (Table~\ref{tab:arch_ablation}).
The bilinear formulation consistently achieves the strongest overall performance, particularly on in-domain datasets, while the patch-specific variant performs second-best in most settings. The shared linear formulation remains competitive on the out-of-domain CCNP samples, consistent with the idea that reduced parameterization benefits generalization.
Overall, the proposed bilinear strategy achieves the strongest predictive performance, supporting the effectiveness of combining constrained parameterization with an explicit network-level inductive bias. 

\textbf{Network Structure and Parcellation.}
To evaluate the interaction between network-based patching and the proposed network-specific bilinear tokenization, we design controlled ablations that modify patch definitions while keeping all other architectural and training components fixed, isolating the effect of anatomically grounded tokenization in NERVE.
First, we perform 100 permutation baselines by randomly shuffling region indices prior to patch assignment within the 17-network parcellation~\cite{ThomasYeo2011TheConnectivity}.
This serves as a null model without meaningful network structure.
Second, we evaluate a vanilla MAE configuration that partitions the FC matrix into fixed continuous $16\times16$ square patches following image-based practice~\cite{He2021MaskedLearners}. We then encode these patches using the \textit{shared}, patch-\textit{specific}, and proposed \textit{bilinear} network-aware formulation, thereby removing explicit functional network grounding.
Third, we test a coarser 7-network parcellation~\cite{ThomasYeo2011TheConnectivity} to assess sensitivity to network granularity.
Table~\ref{tab:parcellation_ablation} reports performance and one-sided permutation test p-values computed against 100 region-permuted baselines. NERVE with the 17-network parcellation achieves the highest performance and significantly outperforms the baselines, supporting the relevance of anatomically informed patch definitions. Notably, when network structure is disrupted, patch-specific linear tokenization tends to outperform the bilinear formulation, suggesting that the proposed bilinear factorization is most effective when aligned with meaningful functional organization.
\section{Discussion \& Conclusion}
\label{sec:conclu}
    
Our results consistently demonstrate that incorporating network-level structure improves stability and cross-cohort generalizability of FC representations. Nevertheless, several limitations should be acknowledged. First, evaluation was conducted within developmental cohorts and psychopathology measures. While this setting reflects a clinically relevant and challenging prediction task, further evaluation across additional populations and phenotypic domains will help assess the generalizability of the learned representations.
Second, our framework currently operates on static FC derived from a predefined parcellation. Extensions incorporating dynamic FC patterns~\cite{Hutchison2013DynamicInterpretations} or multimodal integration with structural connectivity derived from diffusion MRI~\cite{Litwinczuk2022CombinationFunction} represent promising directions for extending the proposed framework.
Future work will focus on scaling to more heterogeneous datasets, exploring alternative downstream tasks, and improving interpretability of the learned representations. In particular, future analyses will examine the learned network-specific weights in the bilinear embedding and characterize attention patterns between network-aware tokens to better understand how large-scale network connectivity patterns across parcellations contribute to representation learning.

In this work, we introduced a self-supervised framework that incorporates large-scale functional network structure into brain connectivity representation learning through a bilinear tokenization strategy. Our results demonstrate that network-aware modeling provides a principled inductive bias for functional connectomics and supports scalable cross-cohort brain–behavior mapping.

    

\begin{credits}
\subsubsection{\ackname} 
This work was supported in part by NIH Grant AA028840 (QZ), a BBRF Young Investigator Grant (QZ), and a NIARR Pilot Resource Grant (QZ).

\subsubsection{\discintname}
The authors have no competing interests to declare that are relevant to the content of this article.
\end{credits}

%
%
%
%


\end{document}